# SAML-QC: a Stochastic Assessment and Machine Learning based QC technique for Industrial Printing

Azhar Hussain
College of Information and Communication Engineering, Harbin Engineering University, 150001, Harbin, China
engrazr@hrbeu.edu.cn

***Abstract*— Recently, the advancement in industrial automation and high-speed printing has raised numerous challenges related to the printing quality inspection of final products. This paper proposes a machine vision based technique to assess the printing quality of text on industrial objects. The assessment is based on three quality defects such as text misalignment, varying printing shades, and misprinted text. The proposed scheme performs the quality inspection through stochastic assessment technique based on the second-order statistics of printing. First: the text-containing area on printed product is identified through image processing techniques. Second: the alignment testing of the identified text-containing area is performed. Third: optical character recognition is performed to divide the text into different small boxes and only the intensity value of each text-containing box is taken as a random variable and second-order statistics are estimated to determine the varying printing defects in the text under one, two and three sigma thresholds. Fourth: the K-Nearest Neighbors based supervised machine learning is performed to provide the stochastic process for misprinted text detection. Finally, the technique is deployed on an industrial image for the printing quality assessment with varying values of n and m. The results have shown that the proposed SAML-QC technique can perform real-time automated inspection for industrial printing.**

## I. Introduction

The quality control and inspection of the high speed printing process is a critical topic in various industries ranging from electronics to pharmaceutical products. The advent of higher quality requirements from end-users and higher costs of raw materials has limited the profit margin of labeling producers. The label producers need to improve quality of their products and increase the production efficiency by keeping the cost and management thresholds to a minimum level. Typical industrial label printing mechanisms include offset-printing, letterpress-printing, combination-printing and Felxo-Printing. Various challenges faced by the labels are the limited sized of text-containing region, relatively small size of labels, clear text, ink-flow control for same shades and text-alignment.

Usually the common defects encountered in printing process are misprinting, text fading and various shades in printed text. According to the report [1], in 2010 the Chinese pharmaceutical label market demand was more than 165 million m², and the increase is predicted to be 10% to 12% in the next three to five years. Europe and America had a market demand of 300 million m² and 250 million m² respectively and both are expected to increase by about 3% annually. It is evident that the market demand of pharmaceutical labels in western countries is higher than in China.

Chinese market is expected to witness an increase in demand in the coming years. It is also reported that in the past three years, the labor costs have increased dramatically and QC management has become more difficult. Manual QC inspection can be easily affected by the factors such as lack of experience, emotions, human eye perception, and environmental conditions. These factors lead to an inconsistent and poor standard of inspection. Although a review process is usually added in the inspection process in order to recheck the misprinted labels and quality defects, yet it cannot guarantee 100% quality inspection and also leads to extra labor costs and time consumption. In order to win the customer satisfaction the inspection QC should be improved through automation. Even tiny mistakes in the label printing can put a bad impression in the overall product quality and can reduce a good customer relationship. Therefore, automatic inspection system is an obvious choice for the industries and it will have room to grow in the near future.

The proposed scheme performs quality inspection in three steps. The first step is related to the inspection of printed text alignment with respect to the industrial object. Although the requirement for printing alignment is company specific yet the proposed mechanism is adaptable and adjustable according to the specific requirement. The second step performs quality inspection based on detection of varying printing shades in the detected text. The third step is aided by supervised machine learning and it performs the detection of misprinted text. The rest of the paper is composed as follows. Section II addresses the related work for computer aided quality inspection of printing. Section III explains the SAML-QC algorithm. Section IV shows the results of performing the proposed inspection on a given industrial object. Finally, section V presents conclusion and future work.

## II. Related Work

Recently, a few researches have investigated computer-aided detection and image quality assessment. In [2] a technique based on comparison of an inspected document with its referential version is discussed. In [3] an image quality assessment algorithm is proposed that does not rely on reference images and its general framework emulates human quality assessment by first detecting visual components and then assessing quality against an empirical model for face detection. In the algorithm of Rowley et al. [4] a neural network is trained to detect face patterns in a region of 20-by-20 pixels. The determination of an arbitrary image such that if the square is a face region, the square is down-sampled to the size of 20 by 20 and equalized, resulting in a normalized signal.

A similar proposition holds for the face detector of Viola and Jones [5]. In their algorithm a feature based classifier is trained over a squared box of 24-by-24 pixels, with each feature defined by a feature template composed of a group of rectangular sub-windows. In [6] two machine learning algorithms are used for feature selection: Mixture of Gaussian and Redial Basis Function (RBF). The first one is a statistical distribution estimation based algorithm, while the second is a function approximation based algorithm. However, the preliminary experiment based only on face detection and face based quality modeling yielded encouraging results. However, more work needs to be done in ]areas such as object detection, feature selection and machine learning to better establish this method.

As an efficient alternative, the machine vision systems can filter out the physical limitations and subjective judgmental decisions of humans. In [7] an image processing technique for the development of a low-cost machine vision system is explored for the inspection of the pharmaceutical capsule. This work discusses the two-part gelatin capsule inspection system by using image processing techniques for border tracing and approximation of the capsule to a circle. A quality control feedback performs pass/reject decision and put capsules to the appropriate bin. In [8] a new approach for detecting the printing accuracy based on the technology of machine vision is presented.

The process consists of image acquisition, filtering, segmentation and image matching. In this paper, JSEG [9] algorithm is used to segment the textile printing images with obvious texture; and then make a match for the segmented edge information, to obtain deviation area of the image, and calculate the deviation position. However, in their experiments, several limitations are found for the system. One case is when the similar colors between the two neighbor regions cannot be segmented. The work in [10] introduced the algorithm based on wavelet packet and the regional analysis. It analyzes the significance of the threshold of binarization in defect detection combined with morphology area analysis method to extract specific features of defects and determine the type of defect.

## II. PROPOSED SAML-QC SCHEME

The block diagram of the proposed scheme is shown in Fig. 1. It shows that the algorithm receives the input image and performs pre-filtering. The pre-filtering process starts by RGB to gray scale conversion and resizing the input image to a width of 500 pixels without losing the aspect ratio. A Gaussian blurr (of kernel size 9×11) filters the salt and pepper noise in the image. Then the Histogram equalization process enhances the contrast. The next step is the initialization of structuring kernels. These kernels play a major role in the morphological closing operation. This operation combines the text-containing region and supports the detection of text containing region for the alignment testing. The next step is the initialization of TopHat morphology [11] to find the white regions against the darker ones.

It follows the computation of Scharr Gradient [12] of the TopHat image for edge detection. The proposed scheme used the sobel operator in horizontal axis to calculate the absolute value element-wise. The minimum and maximum values of Scharr gradient is obtained followed by scaling to the range 0 to 255 per pixel value. The next step performs the morphological closing operation to fill the gaps. It is followed by Otsu’s Auto-thresholding [13] to binaries the image. The morphological closing and dilation process is performed to convert the text containing-region as a combined blob. It is intuitive that this combined block contains the actual area of the text and makes the image ready for the detection of contours. The purpose of finding contours in the image is to detect the text-containing region inside the given object. After that, the text containing region is cropped and stored in a buffer for later use in stochastic assessment. The text containing region is identified by selecting the bounding box with optimum aspect ratio in the process of iterating over the found contours. The red rectangle is displayed on the image of the selected text. The coordinates of the text containing region are used for the testing of text alignment.

### A. *Text alignment assessment*

In printing industry the common text alignment errors are related to the horizontal and vertical alignments. In the current paper it is assumed that the best position of the text is right at the center of the object. Therefore, any text region that is printed too much horizontally or vertically should be identified. However, the height and width of the industrial objects and their respective printed texts are different due to the varying nature of performance of the printing labelers as discussed in [10]. Therefore, proposed algorithm presents a mathematical formulation to detect alignment error of the text inside an industrial object. Fig. 2 shows an abstract image of an industrial object (red portion) with markers showing the dimensions. The width and height of the industrial object are represented by $Wo$ and $Ho$ respectively. The width and height of text containing region are represented by $w$ and $h$. Let $(xo,yo)$ and $(xt,yt)$ represent the top left corners for the object and text boxes respectively as shown in Fig. 2. The absolute difference in terms of pixels between the top edge of industrial object and top of the bounding box of the text region is given by (1), and similarly the absolute differences for down and left are given by (2) and (3).

$$U = abs\{(yo - yt)\} \tag{1}$$

$$D = abs\{(yo + Ho) - (yt + h)\} \tag{2}$$

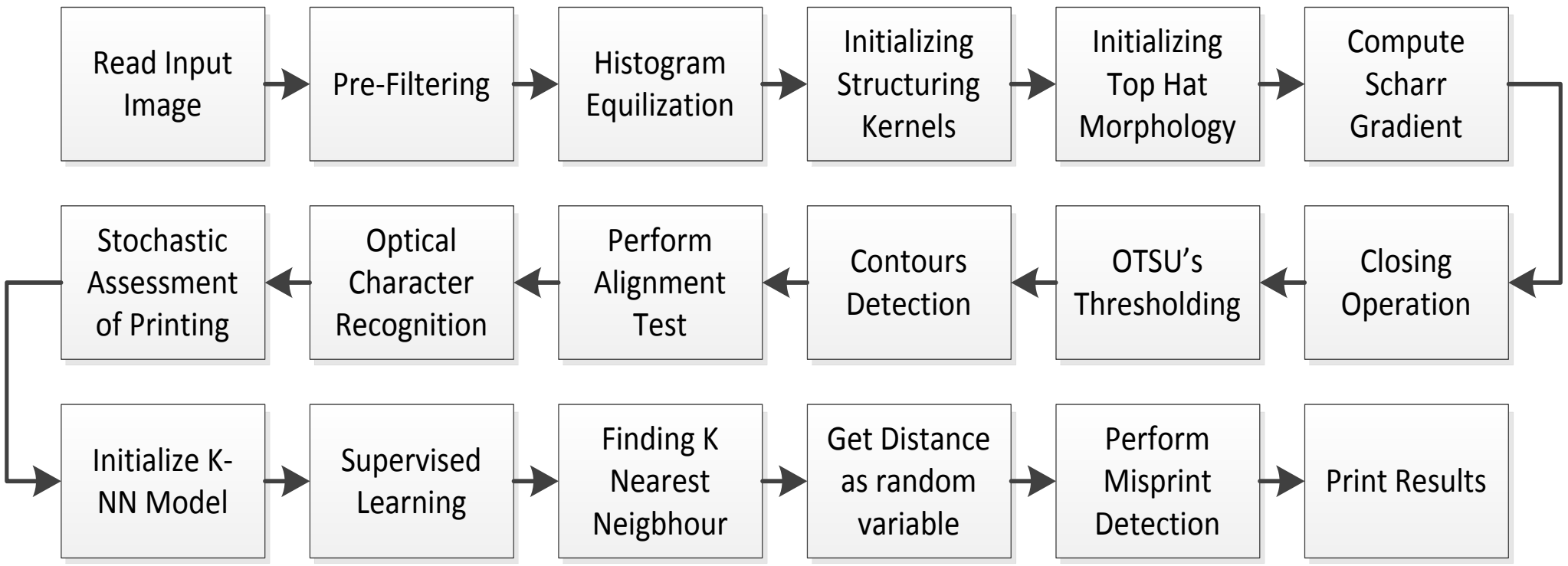

Fig. 1 Block diagram of SAML-QC Algorithm

$$L = abs\{(xo - xt)\} \tag{3}$$

The proposed criterion for text alignment is based on threshold value for vertical and horizontal alignments. Let $UD_{Thresh}$ represents threshold for Up-Down or vertical alignment testing. The value of $UD_{Thresh}$ is set as $h$. Let $L_{Thresh}$ represents the Left-Right or horizontal alignment.

The value of $L_{Thresh}$ is set as Wo/4. It means that it contains the width of one fourth of the full width of object. The decision boundaries for horizontal and vertical alignment passed or failed are given as follows. If $abs\{(U\text{-}D)\} \leq UD_{Thresh}$ , then the vertical alignment is considered as passed else it is failed. If $L \leq L_{Thresh}$ then the horizontal alignment is considered as passed, else it is considered as failed.

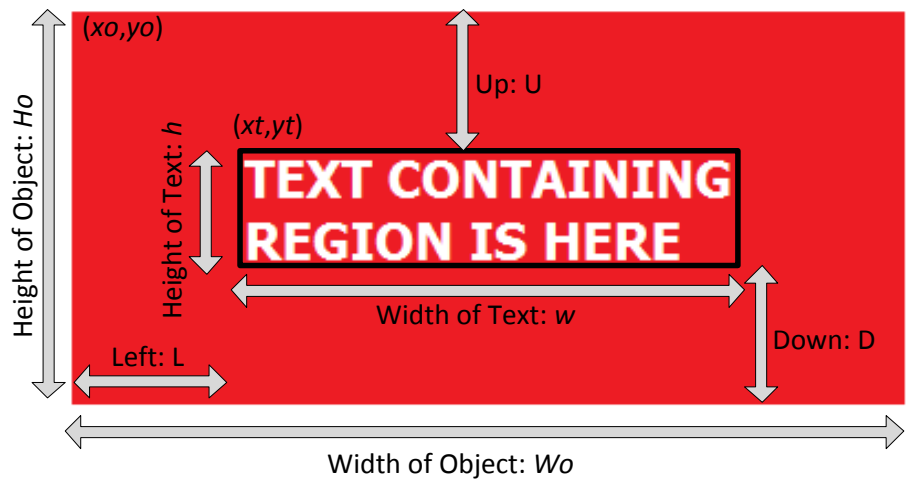

Fig. 2 Parameters for the alignment test of text containing region

## B. *Optical Character Recognition*

After the assessment of alignment the next step is to perform the optical character recognition. For this purpose the proposed scheme crops the text containing region ($xt,yt, x+h,w+h$) of the input image. The median filter is applied to remove any salt paper noise in the image. The next step is to perform the Gaussian blur with emphasis in the vertical direction by selecting the kernel size of 1×5. The reason behind this step is to exploit the nature of text printing. For example in Fig. 2 the vertical difference between the "TEXT CONTAINING" and "REGION IS HERE" is more than the horizontal difference between consecutive individual letters. The next step in the process of obtaining optimum image ready to perform Optical Character Recognition (OCR) is to perform the image contrast enhancement.

The proposed scheme used Bilateral filtering [14] to improve the text regions while suppressing the background. The Bilateral filtering alters the intensity value of each pixel with the weighted average of its neighbors that is Gaussian distributed. The next step is to perform the gray scale conversion and perform Otsu's autothresholding and bit-wise logical NOT operation to obtain binary image. This image is saved as a PNG format. This PNG image is provided as input image to the state-of-the-art Tesseract Open Source OCR Engine v3.02 [20]. The purpose of performing OCR is to get bounding boxes and location of anything that looks like a character. As a result an html file containing the position and sizes of each detected letter is generated. Fig. 3(a)-(b) shows an image that is a cropped and processed by the stated procedure to get the detected boxes of letters in the text and marked with green rectangles.

## C. *Stochastic Assessment of Printing Quality*

The next block performs the stochastic assessment of the printing quality in terms of detection of overly/faded printed text due to the imperfections in printing process as discussed in [1].

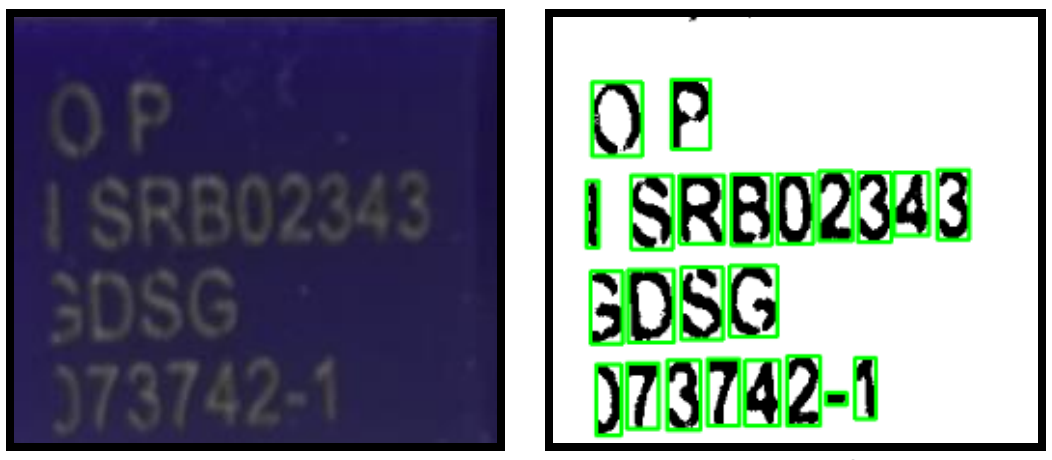

(a) Text area selected (b) Detection of letters

Fig. 3 The detection of bounding boxes of letters

The image in Fig. 3(a) is cropped and selected by the proposed scheme. The intensity level of each letter is varying, yet there is no reference level available for the quality assessment in terms of the intensity levels of characters. Because the distribution of intensity level for each individual letter inside a green box is unknown in Fig. 3(b), so it is assumed that it follows the Gaussian distribution.

### 1) *Detection of printing in varying shades*

The intensity value of each printed pixel can be considered as a random variable $H$ sampled at time $t$, and since its distribution is also unknown therefore, it follows Gaussian distribution. From [15] it is clear that a sequence of random variables is independent and identically distributed (IID), if every random variable has the identical probability distribution and all are mutually independent. Thus it is also assumed that $H$ is IID Gaussian random variable. The noise related to the intensity variation is Gaussian in nature, hence the mean and the acquired image also has additive Gaussian noise $n_i$ with $H$. In order to estimate the parameter $H$, from parametric space to the estimation space $H'$ via observation space $y(H,t)$ at any time $t$, the Maximum Aposteriori Probability (MAP) estimation [16] is used. MAP maximizes the a-posteriori probability, which means most likely value of $H$ is given by (4).

$$\max_{\{H\}} f(H \mid y) = \max_{\{H\}} \left\{ \frac{f(y \mid H) f(H)}{f(y)} \right\} \tag{4}$$

It is quite clear that the probability density function (p.d.f) of parameter $H$ needs to be determined in order to maximize the expression (4). It is assumed that $n_i$ is IID with $\mathcal{N}(0, \sigma_n^2)$ and $H$ is independent of $n_i$ with $\mathcal{N}(0, \sigma_H^2)$. Let $K$ numbers of pixels are available for a given box to estimate $H$. The conditional p.d.f of $y$ given $H$ is given by (5).

$$f(y \mid H) = \prod_{i=1}^{K} \frac{1}{\sqrt{2\pi\sigma_n^2}} e^{-\frac{(y_i - H)^2}{2\sigma_n^2}} \tag{5}$$

And the p.d.f of $H$ is given by (6)

$$f(H) = \frac{1}{\sqrt{2\pi\sigma_H^2}} e^{-\frac{(H)^2}{2\sigma_H^2}} \tag{6}$$

As it is known that the conditional p.d.f of $H$ given $y$ is in (7).

$$f(H \mid y) = \left\{ \frac{f(y \mid H) f(H)}{f(y)} \right\} \tag{7}$$

Inserting the values from (5) and (6) in (7) provides the equation (8).

$$f(H \mid y) = \left( \prod_{i=1}^{K} \frac{1}{\sqrt{2\pi\sigma_n^2}} \right) e^{-\sum_{i=1}^{K} \frac{1}{2\sigma_n^2}(y_i - H)^2} e^{-\left(H^2/2\sigma_H^2\right)} \frac{1}{f(y)\sqrt{2\pi}\sigma_H} \tag{8}$$

The expression in (8) can be defined in terms of $q(y)$ as shown in (9).

$$f(H \mid y) = q(y) e^{-(1/2\sigma^2)(H - \sigma^2/\sigma_n^2 \sum_{i=1}^{K} y_i)^2} \tag{9}$$

The notation $\sigma$ is given by (10).

$$\sigma^2 = \frac{1}{(K\sigma_H^2 + \sigma_n^2)} \sigma_H^2 \sigma_n^2 \tag{10}$$

It should be noted that $q(y)$ is only the function of $y$, therefore, the best estimate $H_{MAP}$ of $H$ is the value where $f(H|y)$ gets the highest peak, which is obtained from (11) when $H=H_{MAP}$.

$$H_{MAP} = \sigma^2 / \sigma_n^2 \sum_{i=1}^{K} y_i \tag{11}$$

Equation (10) can be solved to get (12)

$$\sigma^2 = \frac{\sigma_H^2}{(\sigma_H^2 + \sigma_n^2 / K)} \frac{1}{K} \sum_{i=1}^{K} y_i \tag{12}$$

If $\sigma_S^2 \gg \sigma_n^2 / K$ , then the best estimate of $H$ is given as (13).

$$H_{MAP} \approx \frac{1}{K} \sum_{i=1}^{K} y_i \tag{13}$$

$H_{MAP}$ is the best estimate of intensity levels of the character containing region of a detected box, yet its value varies for every other detected box and its probability distribution is also unknown so it can also be assumed as a random variable that follows Gaussian distribution. If $B$ represents the number of detected character boxes in a given image, then the set of real values $h_{MAP}(u)$ assigned to $H_{MAP}$ for all detected boxes is shown in (14), where $u$ represents the index of each box.

$$H_{MAP} = \{h_{MAP}(1), h_{MAP}(2), h_{MAP}(3), \ldots\ldots., h_{MAP}(\mathrm{B})\} \tag{14}$$

Let E [$H_{MAP}$], be the expectation of $H_{MAP}$ and it is expressed as (15). The variance $\sigma^2_{HMAP}$, can be found by (16)

$$E\ [H_{MAP}] = \sum_{u=1}^{B} h_{MAP}(u) P\{H_{MAP} = h_{MAP}(u)\} \tag{15}$$

$$\sigma^2_{HMAP} = E[(H_{MAP} - E\ [H_{MAP}])^2] \tag{16}$$

*Definition 1*: (**Set of Bad Boxes**) The set of all those boxes such that whose members do not satisfy the conditions in (17) is called set of bad boxes.

Where $n$ represents the quality index and smaller value of $n$ corresponds to higher demand of quality of label printing:

$$\left(E[H_{MAP}] - n\sigma_{HMAP}\right) \le h_{MAP}(u) \le \left(E[H_{MAP}] + n\sigma_{HMAP}\right) \tag{17}$$

The selection of bad-box is performed by (17), as one of the main concerns in the quality inspection of label printing is to find the overly printed or faded printed characters. All the character boxes that fall under the stated condition (17) are considered as members of the set of good boxes. It is quite intuitive that high quality printing demands larger set of good boxes. Therefore, the proposed scheme used two separated counters called $GB_c$ and $BB_c$ representing the numbers of good boxes and bad boxes respectively. The % Quality-Success for similar intensity levels $QS_I$ is given by (18).

$$QS_I = \frac{GB_c}{(BB_c + GB_c)} \times 100 \tag{18}$$

It should also be noted that the value of $QS_I$ has a direct relationship with quality index $n$. The smaller value of quality index means strict requirement of quality and for the same image $QS_I$ decreases with the decrease in value of $n$ and vice versa.

In order to plot the probability density estimate of $H_{MAP}$ the procedure for kernel density estimation [17] is used. It returns a probability density estimate *f* for the sampled data in the vector or two-column matrix. It estimates the density at 100 points for univariate data. Fig. 4 shows the plot of probability density estimate for $H_{MAP}$ of the image in Fig. 3(a). It is clear that p.d.f approximately follows Gaussian curve with a certain value of mean and variances.

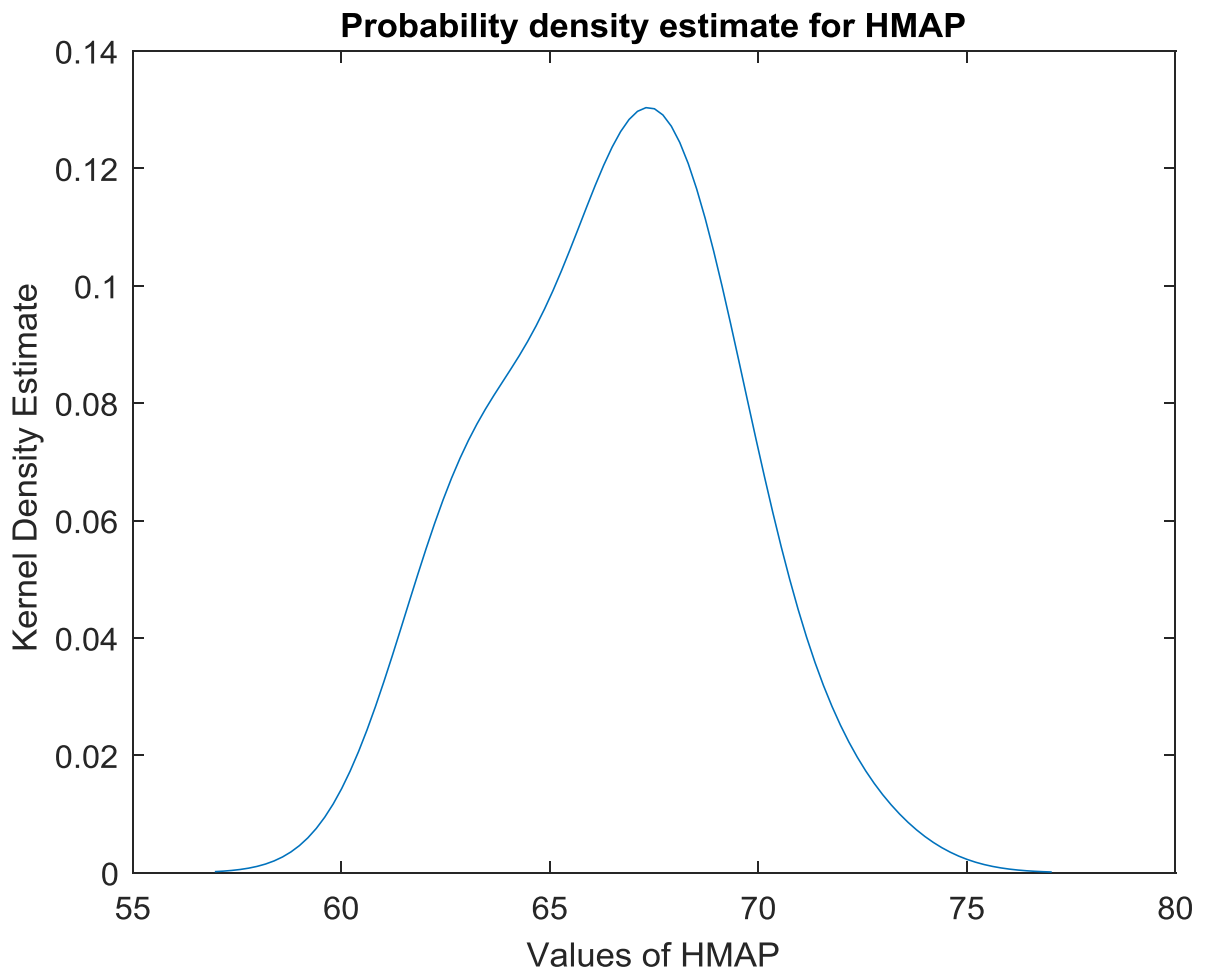

Fig. 4 Kernel density estimate plot of $H_{MAP}$ for the image in Fig. 3(a)

#### 2) *Detection of misprinted boxes*

The detection of misprinted box is challenging in terms of its probability of occurrences and detection. The OCR engine can detected it as a text although being probably wrong detected and classified. The proposed scheme achieves the detection of misprinted box by deploying supervised machine learning using *k*-Nearest Neighbors (*k*-NN). The *k*-NN algorithm is a non-parametric method used for the classification and regression [18] in the pattern recognition. The supervised training aggregates the human responses to the appearance of a letter inside the detected box. The purpose of this step is to provide supervised learning samples and responses for the *K*-NN algorithm.

The procedure for supervised learning in the proposed scheme takes input sample image of industrial object printed with no misprints and the characters are alphabets from A to Z and numeric from 0 to 9 and a -. Let call the set of these input images as no-misprinted-images (NMI). The training process starts by iterating the NMI for all detected boxes and for each detected box a human response is provided as a label and aggregated to a human responses database (HRD) file, because the human needs to press the related key from Keyboard. As for each detected letter box its ratio width/height = 2/3, therefore each box is aggregated as a matrix of 20x30 pixels in a separate matrices received database (MRD) file.

Therefore, at the end of training process two (HRD and MRD) files are ready for the *K*-NN algorithm. The *K*-NN algorithm takes these two files as input and finds the nearest neighbor in terms of its output as Hamming distance *D* as stated above. It is quite intuitive that *D* will have higher value for the misprinted or unknown letters, because the training process did not consider those letters. The value of *D* for each printed letter can be considered as a random variable sampled from a random process at time *t*, and since its distribution is also unknown therefore, it is assumed that it follows Gaussian distribution. Based on the fact stated by [19], the proposed scheme also assumes that *D* is IID Gaussian random variable. The next step is finding the expectation of *D* for each letter box. It should be noted that *D* is normalized by the size of box, i-e. 20x30.

Let us represent E[*D*] and $\sigma_D^2$ as the mean and variance of *D.* And computing these values using the similar way as described in (15) and (16), which provides the second order statistics for the distance. Let the set of real values *d(u)* assigned to the random variable *D* is given by (18) .

$$D = \{d(1), d(2), d(3), \ldots\ldots, d(\mathrm{B})\} \tag{18}$$

*Definition 2*: (**Misprinted Box Detection**) The set of all those boxes such that whose members do not satisfy the conditions (19) is called set of misprinted bad boxes.

$$\left(E[D] - m\sigma_D\right) \le d(u) \le \left(E[D] + m\sigma_D\right) \tag{19}$$

Again the quality index *m* represents the strict boundary for the selection of misprinted boxes. All the character boxes that fall under the stated condition (19) are considered as members of the set of good printed boxes. It is quite intuitive that high quality printing demands larger set of good boxes. Therefore, two separated counters called $GPB_{count}$ and $MPB_{count}$ represent the numbers of good printed boxes and misprinted boxes respectively. The % Quality-Success for good printed boxes $QS_{GPB}$ is given by (20).

$$QS_{GPB} = \frac{GPB_{count}}{(MPB_{count} + GPB_{count})} \times 100 \tag{20}$$

The final step is the printing of the acquired results related to the text alignment, $QS_I$ and $QS_{GPB.}$ These results can not only draw the outcome of printing inspection process but can also provide feedback to the printing labelers to automatically adjust the process according to the type of printing errors.

## III. Results

Table I summarize the values of horizontal and vertical alignments for sample image in Fig. 5. The results support the human observation of central text alignment for this kind of particular object. It should also be noted that text alignment specification and parameters are user specific.

Table I. Text alignment assessment results for the given image

| Images | *U* | *L* | *D* | $UD_{Thresh}$ | $L_{Thresh}$ | *Alignment Testing* | |
|---|---|---|---|---|---|---|---|
| | | | | | | *Horizontal* | *Vertical* |
| 1 | 42 | 59 | 50 | 170 | 119 | Passed | Passed |

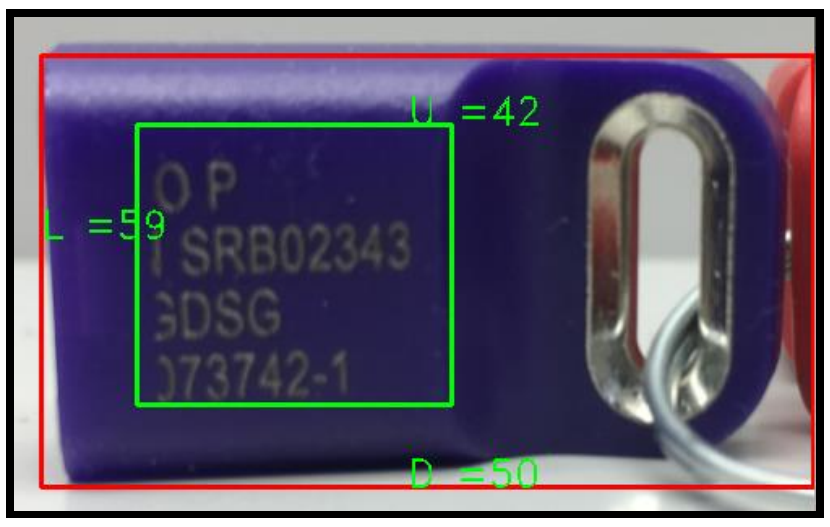

Fig. 5 Text alignment detection

Fig. 6 shows the bar graph of the estimated values for the random variable $H_{MAP}$ for each detected text in Fig. 3(b).

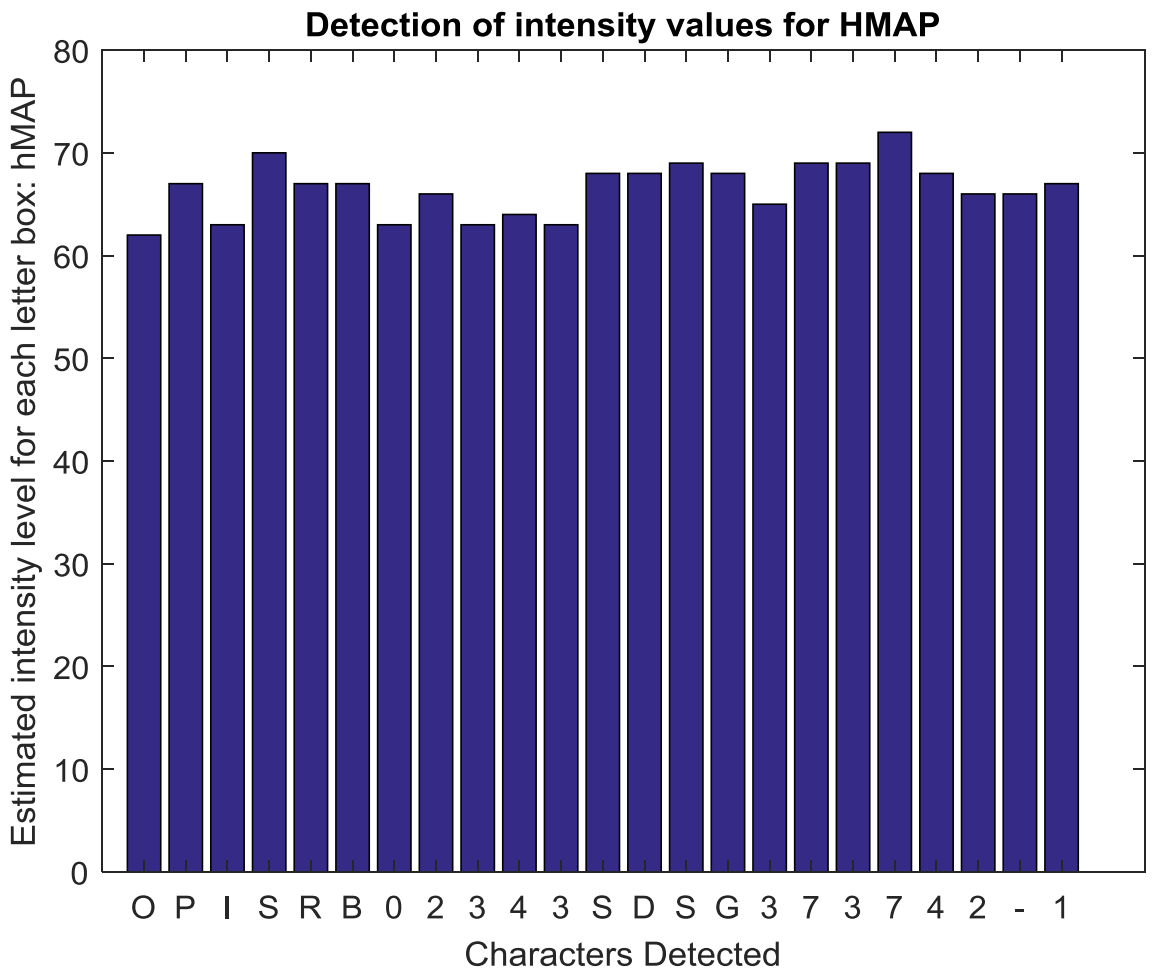

Fig. 6 Estimated intensity values of $H_{MAP}$

Fig. 7 shows the histogram of $H_{MAP}$ in terms of numbers of boxes and the value of random variable $H_{MAP}$. It also shows the corresponding Normal Distribution curve fitted based on the values of $H_{MAP}$.

Table II summarizes the results of printing quality assessment in order to test the intensity variations in the printed labels. It shows mean, variance, Quality (Q) factor $n\sigma_{HMAP}$, the sum of good and bad boxes, and finally the $QS_I$. It is obvious from the results that Q factor decides the value of $QS_I$.

The value of quality index is varied as $n$ = 1, 2, and 3 that increases the value of $GB_c$ as 13, 22 and 23 respectively. Additionally, the values for $QS_I$ for $n$ = 1, 2, and 3 are 56.52%, 95.65% and 100% respectively.

Table II. Printing Quality assessment results for $H_{MAP}$

| Q Index $n$ | Mean $E$ [$H_{MAP}$] | Variance $\sigma^2_{HMAP}$ | Q factor $n\sigma_{HMAP}$ | *Box counts* $GB_c+BB_c$ | $QS_I$ (%) |
|---|---|---|---|---|---|
| 1 | 66 | 6 | 2.44 | 13 + 10 | 56.52 |
| 2 | 66 | 6 | 4.89 | 22 + 1 | 95.65 |
| 3 | 66 | 6 | 7.35 | 23 + 0 | 100 |

Fig. 8 shows the output of SAML-QC scheme for the detection of bad printed labels. The higher quality control requires lower value of $n$.

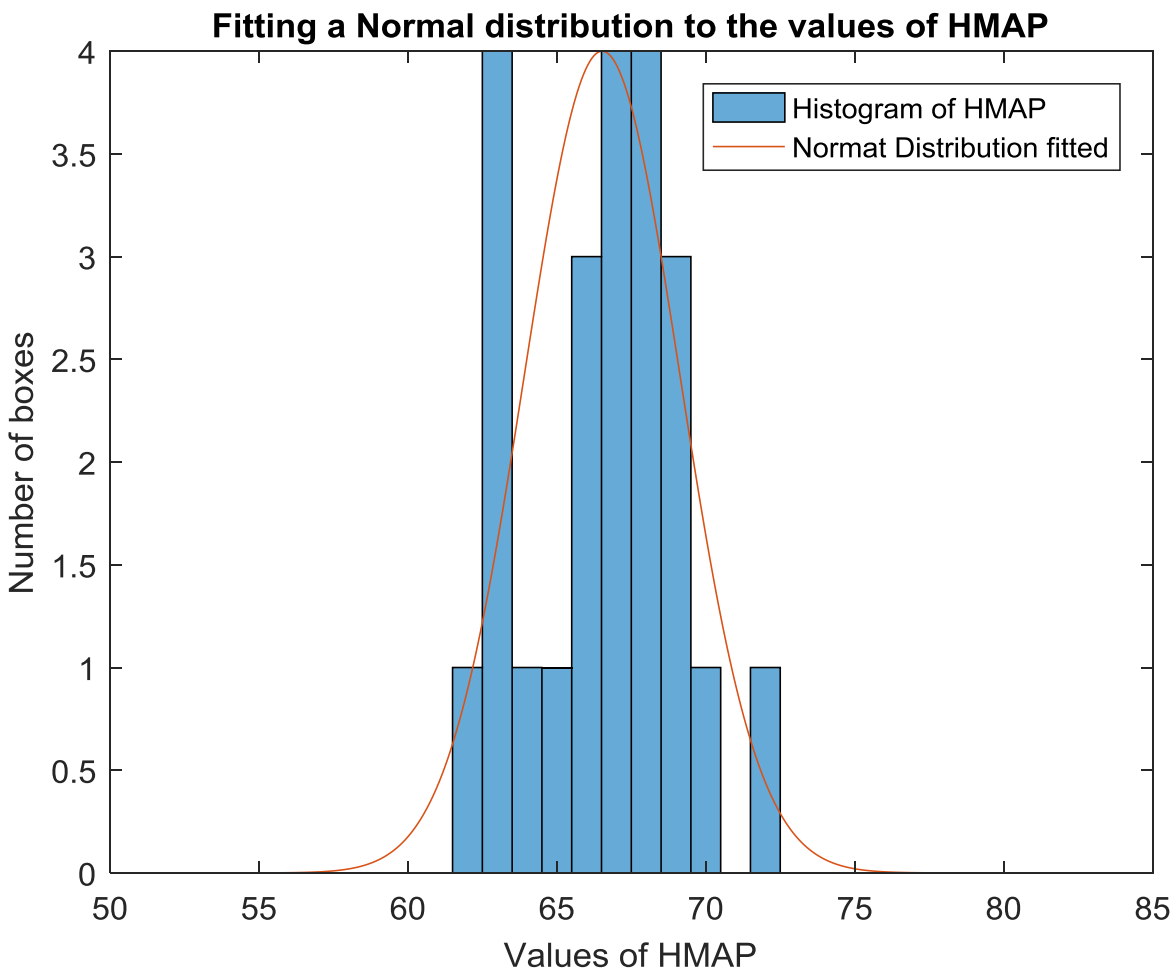

Fig. 7 Plot shows fitting a normal distribution to the values of $H_{MAP}$

Figure 8 (a) shows that for the restriction $n$ = 1, all good boxes are detected and marked with green color, whereas, all the boxes in red color are the ones with value of $h_{MAP}(u) \geq$ E[$H_{MAP}$] + $n\sigma^2_{HMAP}$, whereas, boxes with $h_{MAP}(u) \leq$ E[$H_{MAP}$] - $n\sigma^2_{HMAP}$ are represented by yellow boxes because they belong to the faded printing category. Similarly, Fig. 8(b) and Fig. 8(c) show the output images for $n$ = 2 and 3 respectively.

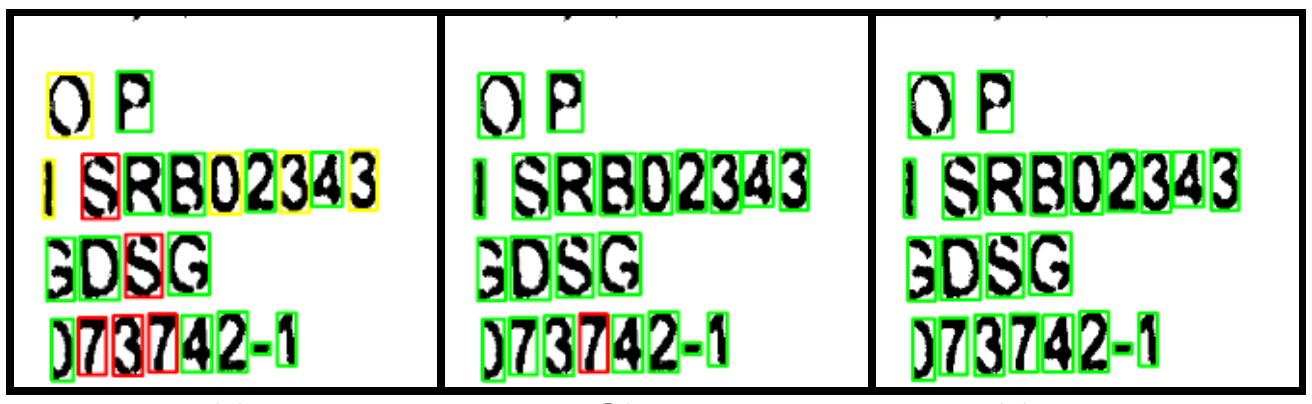

(a) $n$=1 (b) $n$=2 (c) $n$=3

Fig. 8 Detection of bad printed boxes for different $n$

The next assessment is related to the detection of misprinted characters and Fig. 9 shows result of $K$-NN classification performed through supervised machine learning and the distance $D$ for each detected character.

Table III. Misprinted labels assessment results

| Quality Index $m$ | Mean $E$ [D] | Variance $\sigma^2_D$ | Q bound $n\sigma_D$ | *Box counts* $GPB_c+MPB_c$ | $QS_{GBP}$ % |
|---|---|---|---|---|---|
| 1 | 3397.88 | 1.38×10$^7$ | 3727.68 | 21 + 2 | 91.30 |
| 2 | 3397.88 | 1.38×10$^7$ | 7455.36 | 21 + 2 | 91.30 |
| 3 | 3397.88 | 1.38×10$^7$ | 11183.04 | 21 + 2 | 91.30 |

It is clear from the results in Fig. 9 that the two higher spikes for the detected character S and 3 are located at the position of misprinted characters.

The Table III shows results of the process for the detection of misprinted boxes as discussed in section III. It is observed that the proposed scheme put a stable restriction on the misprinted labels and it is shown in Table III that the value of $QS_{GBP}$ remains 91.30% for all three values of $m$.

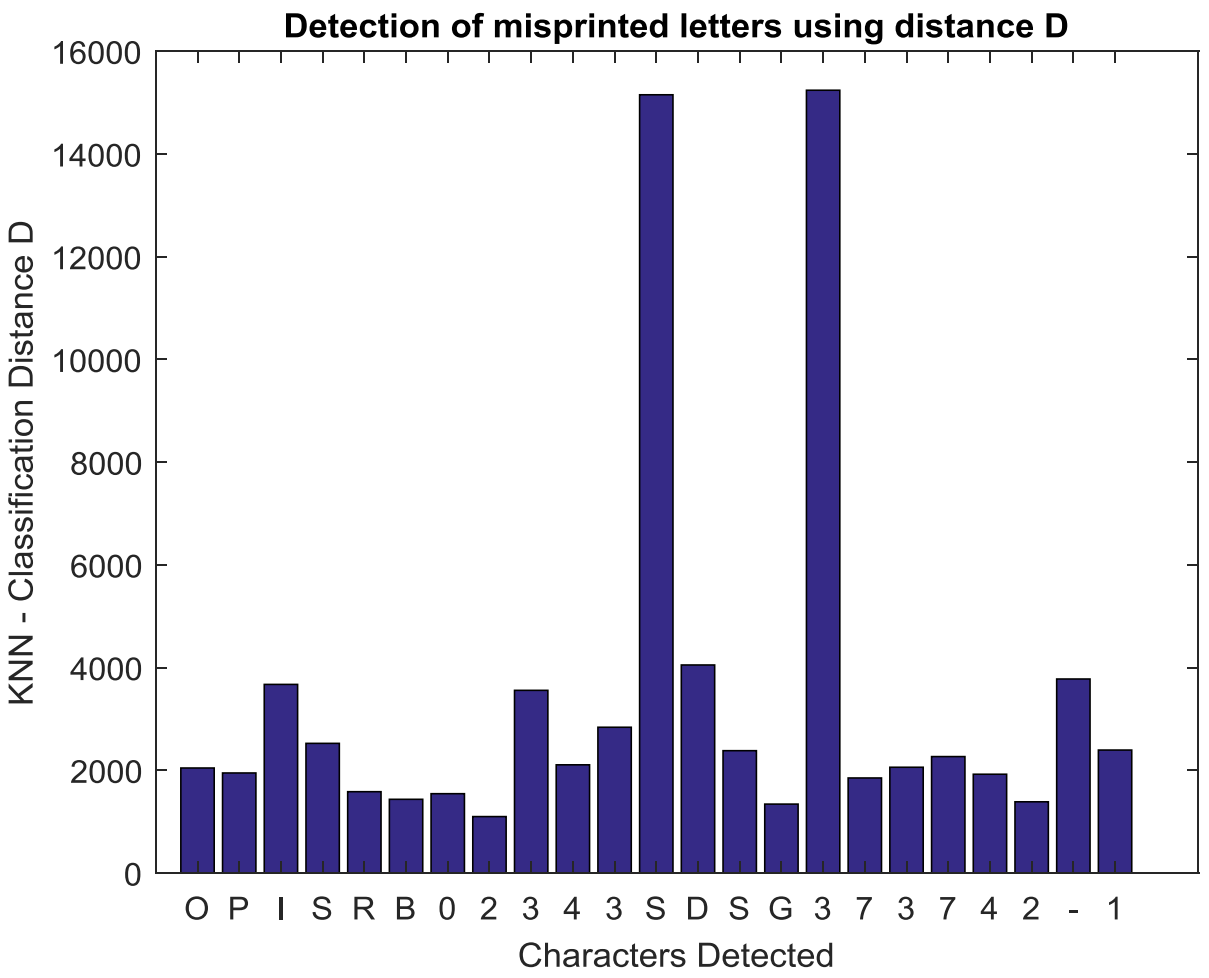

Fig. 9 Bar graph of *K*-NN classification distance *D*

Fig. 10 shows the detection of misprinted labels and are marked with the red boxes for *m=1* , *2* and *3*.

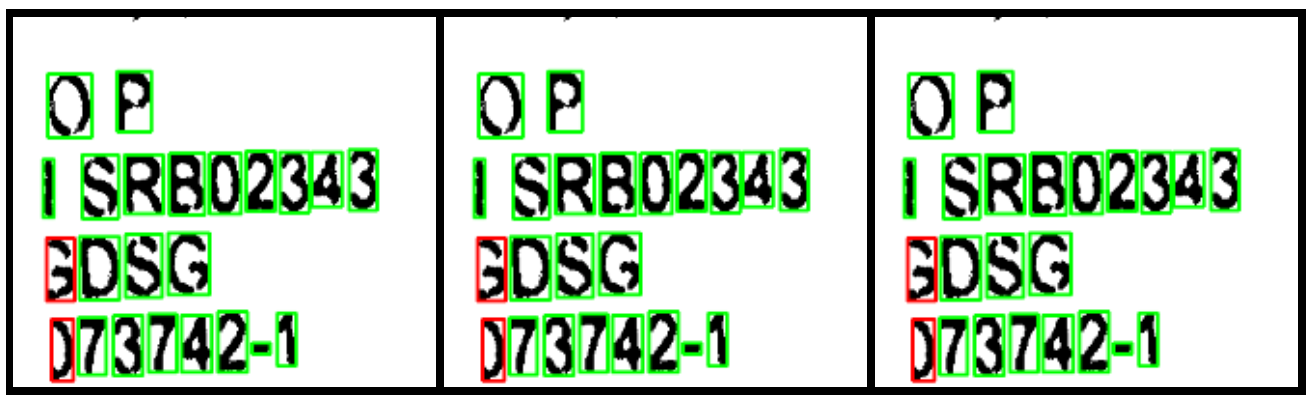

(a) *m*=1 (b) *m*=2 (c) *m*=3

Fig. 10 Detection of misprinted boxes.

## IV. CONCLUSION

This paper proposes a stochastic assessment and machine vision based technique to assess the printing quality of text on industrial objects. The assessment is based on three quality defects such text misalignment, varying printing shades, and misprinted text. It is concluded from the results that second order statistics related to the intensity values of pixels of text located under the threshold region of the original image can provide enough information to perform quality inspection. On the other hand the hamming distance acquired by the K-NN supervised machine learning can also be taken as a random variable and again the second order statistics are helpful in order to detect the misprinted letter. In the future SAML-QC technique needs to be evaluated on various other images such as electronics parts and medical parts.

## ACKNOWLEDGMENT

My deepest acknowledgement is to Prof. Tao JIANG of College of Information and Communication Engineering, Harbin Engineering University for guidance, kind support and wisdom.